\pdfoutput=1

\documentclass[11pt]{article}

\usepackage{acl}

\usepackage{times}
\usepackage{latexsym}

\usepackage[T1]{fontenc}

\usepackage[utf8]{inputenc}

\usepackage{microtype}

\title{Learning-by-Narrating: \\Narrative Pre-Training for Zero-Shot Dialogue Comprehension}

\author{Chao Zhao$^{1}$\thanks{\quad Work was done during the internship at Tencent AI lab.}  \qquad Wenlin Yao$^{2}$ \qquad Dian Yu$^{2}$ \\ \bf
Kaiqiang Song$^{2}$ \qquad Dong Yu$^{2}$ \qquad Jianshu Chen$^{2}$ \\
\texttt{zhaochao@cs.unc.edu}\\ \texttt{\{wenlinyao,yudian,riversong,dyu,jianshuchen\}@tencent.com}\\
$^{1}$ UNC Chapel Hill, Chapel Hill, NC \qquad $^{2}$ Tencent AI Lab, Bellevue, WA}

\usepackage{booktabs}
\usepackage{mathtools}
\usepackage{amsfonts}
\usepackage{color}
\usepackage{tablefootnote}
\usepackage{booktabs}
\usepackage[inline,ignoremode]{trackchanges}

\usepackage{multirow}
\usepackage{graphicx}
\usepackage{xcolor,colortbl}
\definecolor{Gray}{gray}{0.85}
\usepackage{enumitem}

\addeditor{jianshu}

\begin{document}
\maketitle
\begin{abstract}
Comprehending a dialogue requires a model to capture diverse kinds of key information in the utterances, which are either scattered around or implicitly implied in different turns of conversations. Therefore, dialogue comprehension requires diverse capabilities such as paraphrasing, summarizing, and commonsense reasoning. 
Towards the objective of pre-training a zero-shot dialogue comprehension model, we develop a novel narrative-guided pre-training strategy that \emph{learns by narrating} the key information from a dialogue input. However, the dialogue-narrative parallel corpus for such a pre-training strategy is currently unavailable. For this reason, we first construct a dialogue-narrative parallel corpus by automatically aligning movie subtitles and their synopses. We then pre-train a BART model on the data and evaluate its performance on four dialogue-based tasks that require comprehension. Experimental results show that our model not only achieves superior zero-shot performance but also exhibits stronger fine-grained dialogue comprehension capabilities. The data and code are available at \url{https://github.com/zhaochaocs/Diana}.

\end{abstract}

\section{Introduction}

Dialogue comprehension~\cite{sun-2019-dream,cui-2020-mutual} aims to capture diverse kinds of key information in utterances, which are either scattered around or implicitly implied in different turns of conversations. Therefore, it requires different capabilities such as paraphrasing \cite{falke-etal-2020-leveraging}, summarizing \cite{gliwa2019samsum}, and commonsense reasoning \cite{arabshahi2021conversational}.  Recent advances in pre-trained language models (PLMs)~\cite{devlin-2019-bert,radford2019language} have been applied to the problem~\cite{jin2019mmm,liu2021graph}. However, these PLMs are generally pre-trained on formal-written texts, which are different from dialogue data in nature. Specifically, dialogues are composed of colloquial languages from multi-speakers, and utterances usually have complex discourse structures~\cite{afantenos-2015-discourse}. Therefore, applying these models directly to dialogue comprehension, especially in low-resource settings, is sub-optimal.  

\begin{figure*}[t]
    \centering
    \includegraphics[width=6.2in]{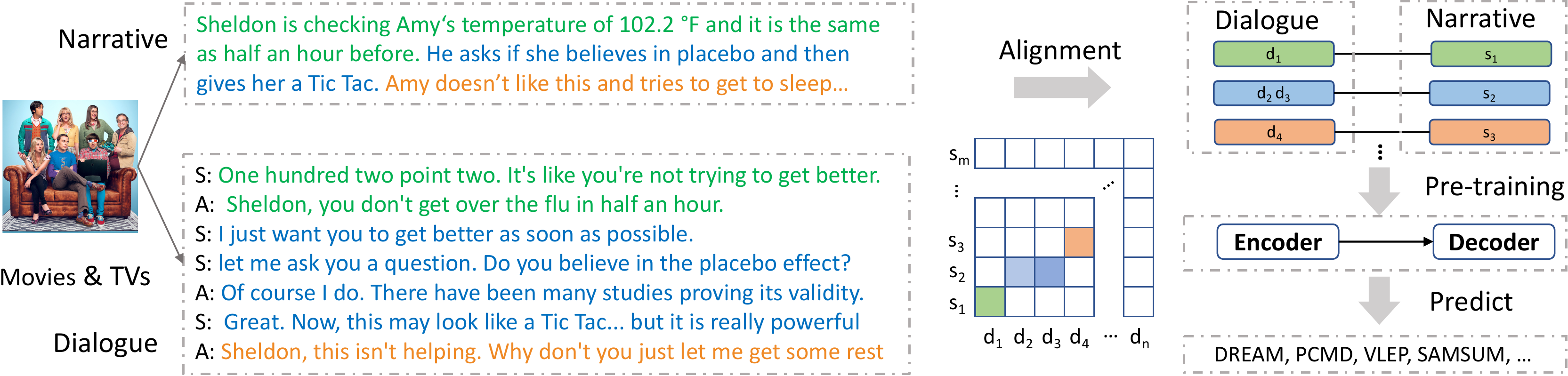}
    \caption{Overview of the \emph{learning-by-narrating} strategy for pre-training a zero-shot dialogue comprehension model (with an encoder-decoder architecture). }
    \label{fig:pipeline}
\end{figure*}

To learn better dialogue representations, recent studies have designed several dialogue-specific pre-training objectives such as speaker prediction~\cite{qiu-2021-different}, utterance prediction~\cite{chapuis2020hierarchical}, response selection~\cite{wu-etal-2020-tod}, and turn order restoration~\cite{zhang-zhao-2021-structural}. These methods, albeit improve over the vanilla PLMs, usually rely on surface-level dialogue information. In particular, they still fail to train the models to explicitly learn the aforementioned capabilities which are critical for dialogue comprehension  (e.g., linguistic knowledge, world knowledge, and commonsense knowledge). Furthermore, it was not able to incorporate knowledge beyond dialogue (e.g., non-verbal communications between speakers, as well as time and location information), which are also crucial for dialogue comprehension.

To pre-train a zero-shot dialogue comprehension model with the aforementioned features, we develop a novel generative pre-training strategy that \emph{learns by narrating} the key information from a dialogue input (see Figure \ref{fig:pipeline} for an example). In particular, the generated narrative text is supposed to not only (i) paraphrase the gists of the dialogue but also (ii) carry certain inferred information (e.g., the time and location of a scene and relations between speakers) that are not explicitly mentioned in the dialogues.
Learning to narrate such information helps the model to learn varied lexical, syntactic, and semantic knowledge of dialogue. It also enhances the model's ability to infer extra information beyond the literal meaning within dialogues, which will benefit the model's capability of dialogue comprehension.

However, the \emph{learning-by-narrating} strategy would require a dialogue-narrative parallel corpus, which, to our best knowledge, is not publicly available. For this reason, we first create \textbf{DIANA}, a large-scale dataset with ({\bf{DIA}}logue, {\bf{NA}}rrative) pairs automatically collected from subtitles of movies and their corresponding plot synopses. We consider dialogues from movie subtitles as they are close to daily human-to-human conversations~\cite{zhang2019automatically}. In addition, the movie synopses include rich narrative information, which is helpful for dialogue comprehension. After data collection and strict quality control, we obtain a dataset with $243$K (dialogue, narrative) pairs written in English. As the automatic data construction procedure is language-independent, it can be applied to low-resource languages as well.

We then pre-train a BART model~\cite{lewis2019bart} on the constructed corpus with the proposed \emph{learning-by-narrating} strategy, and evaluate it on four dialogue-based tasks that require comprehension. In zero-shot settings, our pre-trained model outperforms the BART baseline on all tasks by a large margin (e.g., $+8.3\%$ on DREAM~\cite{sun-2019-dream}), demonstrating the success of our approach. 

The contributions of this paper are three-fold:
\begin{itemize}
    \item We propose a novel \emph{learning-by-narrating} pre-training strategy for dialogue comprehension;
    \item We release \textbf{DIANA}, a new large-scale dialogue-narrative parallel corpus;
    \item Experiments show that our pre-trained dialogue comprehension model achieves superior zero-shot performance on a variety of downstream tasks.
\end{itemize}

\section{DIANA: A Dialogue-Narrative Corpus}
\label{sec::data}

In this section, we describe the procedure to create the dialogue-narrative parallel dataset. 

\subsection{Data Collection and Segmentation}

We collect 47,050 English subtitles of movies and TV episodes released from Opensubtitle \cite{lison2018opensubtitles2018} and their corresponding synopses from online resources such as Wikipedia and TMDB. To link the subtitle and synopsis of the same movie or TV episode, we require a subtitle and a synopsis to have the same title and the release year, as well as a high overlap rate ($> 50\%$) on role names.

The subtitle and synopsis of a movie are too long for a PLM. 
To facilitate pre-training, we split both the subtitle and synopsis into smaller segments and align the related segments from each part to shorter (dialogue, narrative) pairs. We split subtitles using the time interval $\delta_T$ between utterances and split a synopsis into sentences. We set $\delta_T=5s$.

\subsection{Data Alignment}
We aim to align the dialogue sessions $\{d_{1}, \ldots , d_{n}\}$ and narrative segments $\{s_{1}, \ldots, s_{m}\}$ with maximum global similarity to form (dialogue, narrative) pairs. For each dialogue session $d_j$, the goal is to find its corresponding narrative segment $s_i$.

Inspired by \cite{tapaswi2015aligning} in which the narrative in a synopsis follows the timeline of a movie or a TV episode, we develop a dynamic time warping method to find the globally optimal alignment score. During aligning, some narrative segments contain information beyond the dialogue, so they cannot be aligned to any dialogue session. We therefore allow our algorithm to skip at most $k$ narrative segments during alignment searching:
\begin{equation}
    \small
    \mathbf{\mathcal{A}}(i, j)=\max_{0 \le k \le K+1} \mathbf{\mathcal{A}}(i-k, j-1) + \mathbf{\mathcal{S}}\left(s_{i}, d_{j}\right),
\end{equation}
where $\mathcal{A}(i,j)$ denotes the optimal alignment score of the first $i$ narrative segments and the first $j$ dialogue sessions. $\mathcal{S}\left(s_{i}, d_{j}\right)$ is the text similarity between $s_{i}$ and $d_{j}$.

\begin{figure}
    \centering
    \includegraphics[width=1\linewidth]{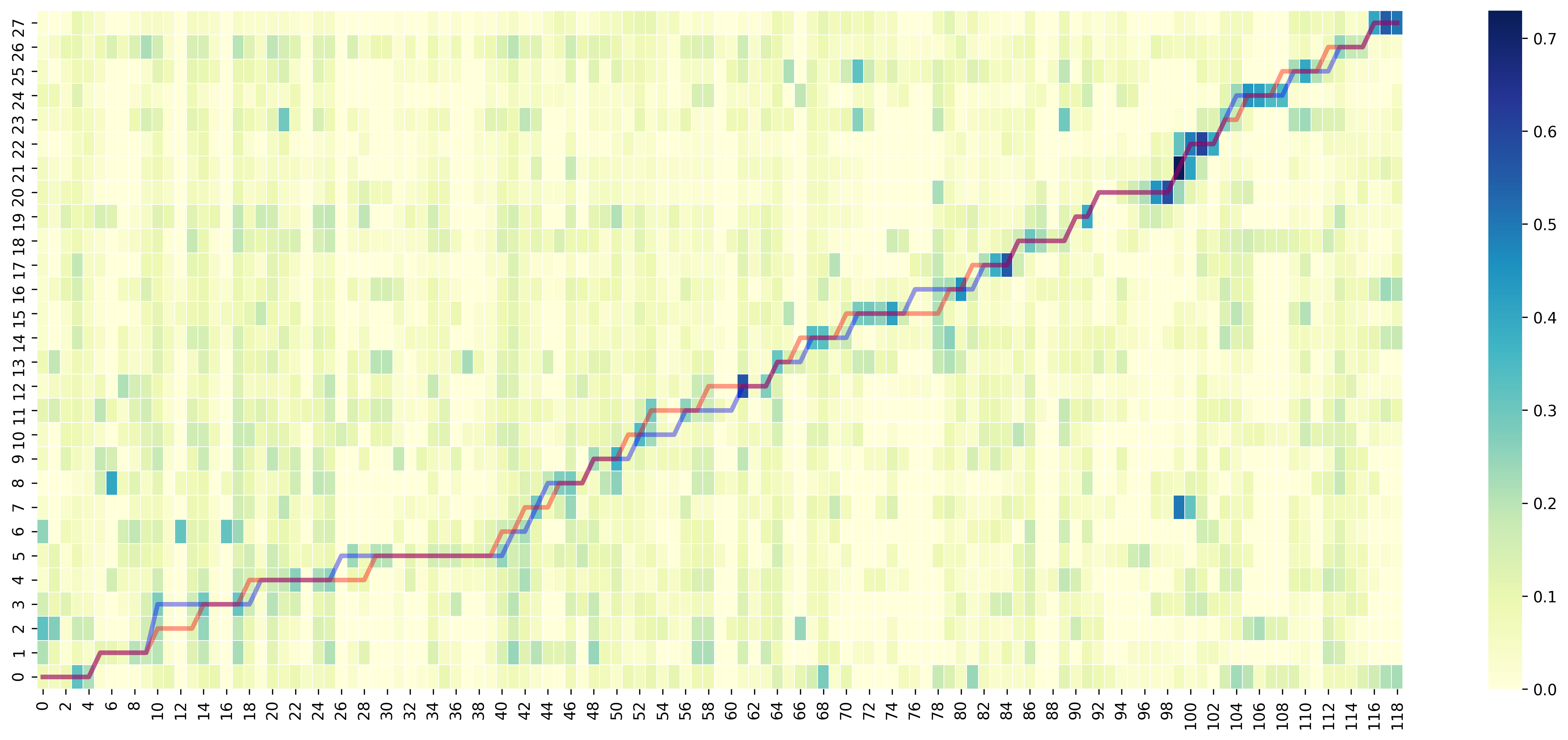}
    \caption{The Alignment of dialogues and narrative segments of a movie. $X$-axis and $Y$-axis are the ID of dialogue sessions and narrative segments, respectively. The variety of colors depicts the different similarity values between a dialogue session and a narrative segment. The blue line is the predicted alignment via normalized TF-IDF while the red line is the gold alignment. }
    \label{fig::align}
\end{figure}

We compare the performance of three text similarity measures: Jaccard similarity, Rouge-1F, and TF-IDF. In consideration of time efficiency, we don't apply more advanced neural methods. We compare these similarity measures on MovieNet dataset \cite{huang2020movienet}, which provides a manual alignment between the segments of subtitles and synopses of 371 movies. \footnote{We use MovieNet for test purposes only.} We evaluate the performance of each similarity measure by alignment accuracy, a.k.a, the percentage of dialogue sessions that are correctly aligned to the corresponding narrative segment. As shown in Table \ref{tab::align_acc}, TF-IDF performs best among all similarity measures. We also find that a narrative-wise $L_2$ normalization of the TF-IDF can further improve the alignment accuracy. It helps to penalize the similarity of $(d_j, s_i)$ when $s_i$ has high similarity with many dialogues (e.g., when $s_i$ contains common words or protagonists' names.) We therefore choose the normalized TF-IDF as our similarity function. We further analyze the errors during alignment and find that 85.94\% of errors happen because the dialogue session is aligned to the previous or next segment of the gold narrative segment. It indicates that most of the errors happen locally. Figure \ref{fig::align} shows an example from MovieNet, where the red line and the blue line indicate the gold alignment and the predicted alignment via normalized TF-IDF, respectively. It shows that the two lines are generally well overlapped except for some local discrepancies.

\begin{table}[t]
		\centering
		\footnotesize
		\setlength\tabcolsep{10pt}
\begin{tabular}{l|c}
 \toprule
 Similarity Function & Accuracy\\
\midrule
Jaccard & $57.98$ \\
Rouge-1F & $60.01$ \\
TF-IDF & $67.20$ \\
TF-IDF normalized & $\mathbf{71.95}$ \\
 \bottomrule
\end{tabular}
		\caption{\label{tab::align_acc}Alignment accuracy of different similarity measures on MovieNet.}
\end{table}

\subsection{Quality Control}
After data alignment, each narrative segment $s_i$ can be aligned to multiple dialogues. To consider the local alignment errors, we also merge the aligned dialogues of $s_{i-1}$ and $s_{i+1}$ to the dialogues of $s_{i}$. Some of these dialogues may not be relevant to $s_{i}$. To select the relevant dialogues, we use a greedy method to incrementally select dialogues until the rouge-F score between the narrative and the selected dialogues doesn't increase. After selection, we concatenate the selected dialogues and preserve their relative position. We finally obtain around 1.5 Million (dialogue, narrative) pairs.

To further improve the quality of data, we filter out pairs where the dialogue and the narrative are irrelevant. To evaluate the relevance, we use two automatic measures: Coverage and Density~\cite{grusky2018newsroom}. Low Coverage and Density indicate that the narrative text is either too abstractive or irrelevant to the dialogue. We thus only select the pairs with $\text{Coverage}>0.5$ and $\text{Density}>1$. After this strict quality control, we obtain 243K (dialogue, narrative) pairs as the final DIANA dataset, which is a high-quality subset of the original dataset. The average length of the dialogue and the narrative are 58 tokens and 18 tokens, respectively.

\subsection{Analysis of Knowledge Type}
\begin{figure}[!t]
    \centering
    \includegraphics[width=0.75\linewidth]{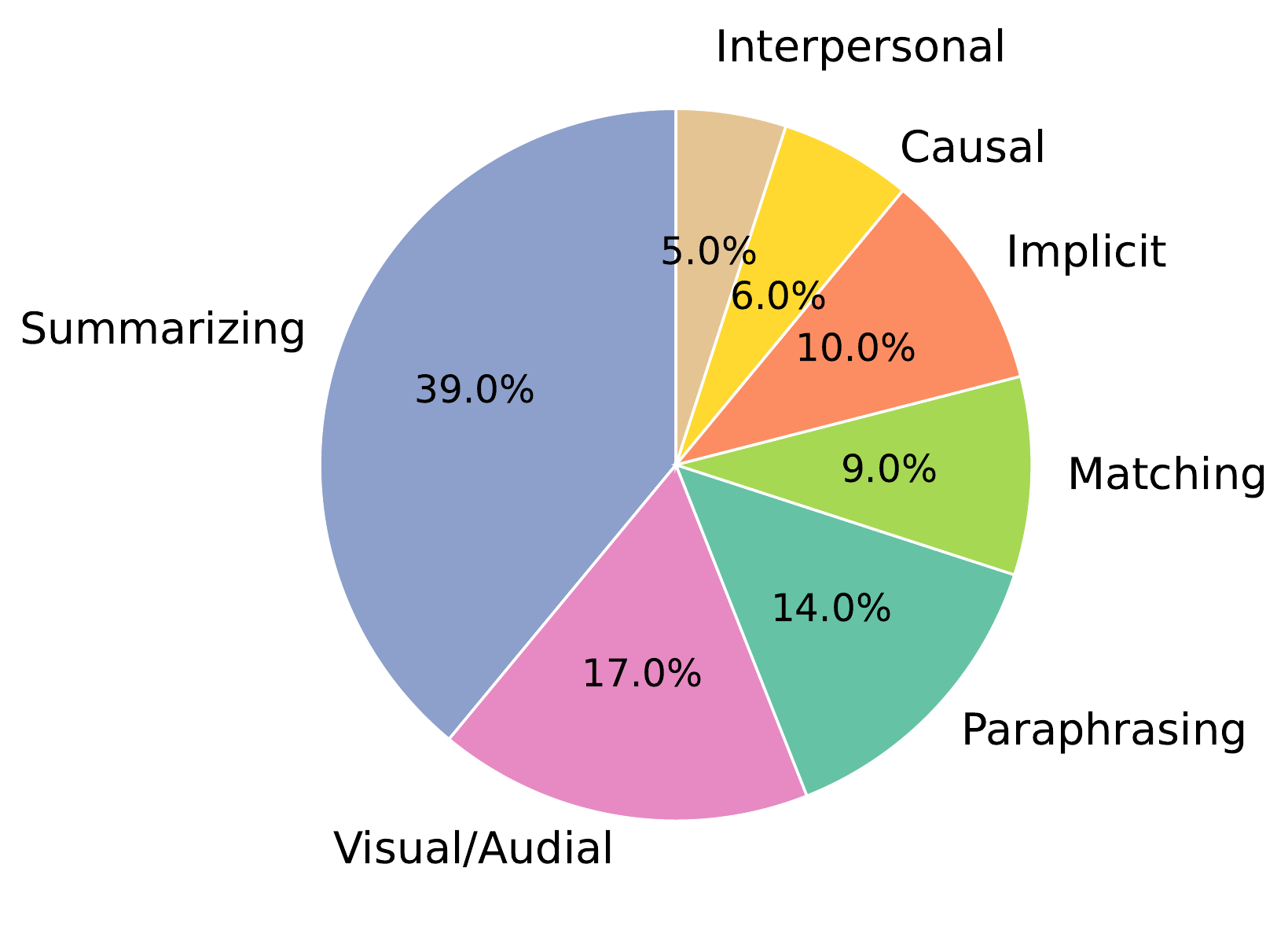}
    \vspace{-0.1in}
    \caption{The knowledge type distribution in DIANA. }
    \label{fig:knowledge_type}
\end{figure}

To analyze what types of knowledge are included in DIANA, we randomly sample 100 instances and manually categorize the relation between dialogue and the corresponding narrative text into seven knowledge types. We show the percentage of each knowledge type in parentheses and in Figure \ref{fig:knowledge_type} as well. The knowledge types are:
\begin{itemize}
    \setlength{\itemsep}{1pt}
  \setlength{\parskip}{1pt}
    \item \textbf{Summarizing} (39\%): The narrative text summarizes multiple utterances as a concise statement to reflect the salient event or information of the dialogue.
    \item \textbf{Visual/Audial} (17\%): The narrative text provides extra visual or audial information of the dialogue, such as the location of the dialogue, the speakers' actions, and ambient sounds.
    \item \textbf{Paraphrasing} (14\%): The narrative text restates speakers' utterances using other words.
    \item \textbf{Text Matching} (9\%): The narrative text is directly copied from the utterances of speakers. 
    \item \textbf{Implicit} (10\%): The narrative text provides extra information that is not explicitly mentioned in the dialogue.
\item \textbf{Causal} (6\%): The narrative text describes the cause and effect relationship between events.
\item \textbf{Interpersonal} (5\%): The narrative text reveals the relationships between speakers.
\end{itemize}

Among these knowledge types, \textit{Summarizing} and \textit{Visual/Audial} are the two most frequent ones. They are followed by
{\it Paraphrasing} and {\it Text Matching}, which contribute to 23\% in total. It also shows that narratives use paraphrasing more often than copying.
Additionally, DIANA contains three higher-level knowledge types that require the awareness of real-world commonsense and more complicated inference such as implicit knowledge, causal relationships, and interpersonal relationships.
The diverse knowledge types in DIANA indicate the benefit of this dataset for dialogue comprehension and other downstream tasks as well. 

\section{Pre-training: Learning-by-Narrating}

During pre-training, we aim to inject the knowledge contained in DIANA into pre-trained models. 
One option is to ask the model to distinguish between a correct narrative and an incorrect narrative via a classification objective. However, it requires carefully designing additional non-trivial negative (dialogue, narrative) pairs. Therefore, we propose to directly generate a narrative text from the given dialogue by maximizing the generative probability:
\begin{equation}
p(\mathbf{y} \mid \mathbf{x} ; \boldsymbol{\theta})=\prod_{t=1}^{|\mathbf{y}|} p\left(y_{t} \mid y_{1: t-1}, \mathbf{x} ; \boldsymbol{\theta}\right),
\label{eq:prob_y}
\end{equation}
where $\mathbf{x}$ are dialogue texts and $\mathbf{y}$ are narrative texts.

There are two main advantages of using the generative objective. First, it can fully leverage the narrative information from each token of the narrative text with no need to construct negative pairs. Second, the pre-trained model can be directly applied to both generative and discriminative downstream tasks without further fine-tuning. For discriminative tasks, we calculate the probability of each candidate according to Equation \ref{eq:prob_y} and choose the most probable candidate as the predicted answer.

\section{Experiments}

In this section, we evaluate the performance of the pre-trained model on four downstream tasks that require dialogue comprehension.

\subsection{Setting}
We use BART, a state-of-the-art sequence-to-sequence model, as our baseline model.\footnote{We also tried T5 and Pegasus in our early experiments but did not observe better performance compared with BART.} We use its released checkpoint and further pre-train the model on DIANA. During pre-training, we concatenate the utterances as the input and update the parameters to maximize the probability of the corresponding narrative. We use Adam as the optimizer, and we set the learning rate and weight decay to $3\!\!\times\!\! 10^{-5}$ and 0.01, respectively. Following previous studies that suggest that a larger batch size helps pre-training, we set the batch size to 1024 and pre-train the model for 1,000 steps.

\subsection{Tasks}
We evaluate our model's ability of dialogue comprehension on four downstream tasks. \textbf{DREAM} \cite{sun-2019-dream} aims to read a dialogue and select the correct answer from options of a dialogue-related question. To make the task similar to our pre-training task, we follow previous work \cite{chen-etal-2021-nli-models} to train a T5 model to convert each (question, answer) pair to a statement. 
\textbf{PCMD} \cite{ma-etal-2018-challenging} is a passage completion task. Given a dialogue and a passage that describes the dialogue, a query is created by replacing a character mention with a variable $x$, and the model needs to recover the character mention. 
\textbf{VLEP} \cite{lei2020more} aims to select the most probable future event given the dialogue of the current event and two candidates of future events. 
\textbf{SAMSum} \cite{gliwa2019samsum} is a dialogue summarization task to create a concise abstractive summary for a dialogue. The first three are discriminative tasks, and SAMSum is a generative task. None of the source dialogues in these tasks are included in DIANA. 

We evaluate the model performance on these tasks under the zero-shot setting. For discriminative tasks, we convert each test instance with $K$ answer candidates as $K$ (dialogue, narrative) pairs. Given the dialogue as input, we evaluate the conditional probability of each narrative according to Equation \ref{eq:prob_y} and choose the most probable narrative as the predicted answer. We use accuracy (ACC) as the evaluation metric for discriminative tasks and ROUGE for the summarization task.

We compare our pre-trained model (Narrator) with strong pre-trained baselines such as GPT-2, RoBERTa, and BART. To investigate the impact of the pre-training objective, we compare with 1) BART-DIAL-DE: the original BART de-noising objectives, which is trained on the dialogue part of DIANA; and 2) BART-CNN-CLS: a classification objective, which is trained using the CNNDM dataset~\cite{see-2017-get} to distinguish between positive and negative summaries based on the documents. Negative summaries are obtained from DocNLI \cite{yin-etal-2021-docnli} by replacing the words, entities, and sentences of positive summaries. We also investigate the quality of DIANA by comparing it with two large summarization datasets: CNNDM and CRD3~\cite{crd3}. We pre-train BART to generate the summaries of these datasets from the corresponding documents and refer to the models as BART-CNN-GEN and BART-CRD3-GEN. Besides the zero-shot models, we list the supervised results finetuned on BART (BART-FT) as a reference for the upper bound.

\begin{table}[t]
		\centering
		\scriptsize
		\setlength\tabcolsep{2pt} %

\centering
 \begin{tabular}{l|c c |c c c c c c} 
 \toprule
 \multirow{2}{*}{} & \multirow{2}{*}{Data} & \multirow{2}{*}{Task} & \multicolumn{1}{c}{DREAM} & \multicolumn{1}{c}{PCMD} & \multicolumn{1}{c}{VLEP} & \multicolumn{3}{c}{SAMSum}  \\
    &&& ACC & ACC & ACC & R1 & R2 & RL \\
 \midrule
 \rowcolor{Gray!93}
BART-FT & - & - & 62.56 & 75.89 & 65.07 & 49.18 & 24.47 & 47.12\\
\hline

GPT-2 & - & - & 41.99 & 45.02 & 54.58 & 10.83 & 0.74 & 11.68 \\
RoBERTa & - & - & 45.22 & 46.25 & 52.28 & - & - & - \\
\hline
\multirow{5}{*}{BART} & - & - & 45.07 & 46.07 & 54.26 & 29.92 & 9.58 & 28.54 \\
 & DIAL & DE & 46.69 & 47.34 & 55.98 & 30.08 & 9.52 & 29.36 \\
 & CNN & CLS & 50.46 & 49.27 & 55.53 & - & - & -\\
\cline{2-9}
 & CNN & GEN & 52.72 & 45.34 & 58.13 & 31.33 & 9.08 & 28.03 \\
 & CRD3 & GEN & 52.96 & 45.71 & 57.12 & 27.07 & 9.09 & 27.64 \\
\hline
\textbf{Narrator} & DIANA & GEN & \textbf{53.41} & \textbf{54.88} & \textbf{58.90} & \textbf{37.27} & \textbf{13.23} & \textbf{36.12}\\
 \bottomrule
 \end{tabular}

		\caption{\label{tab::result_comp}Results on four dialogue-based tasks. For models that require further pre-training, we list the corresponding pre-training dataset and task.}
\end{table}

\subsection{Results}
Results are shown in Table \ref{tab::result_comp}.
Our observations are as follows. (i) When compared with vanilla PLMs, Narrator outperforms GPT-2, RoBERTa, and BART, demonstrating that the learning-by-narrating pre-training objective can improve the model's ability of dialogue comprehension.
(ii) When compared with different pre-training tasks, Narrator outperforms BART-DIAL-DE, and BART-CNN-GEN outperforms BART-CNN-CLS. This indicates that the narrative-guided generative objective is more effective than the de-noising objective and the discriminative objective. (iii) When compared with different pre-training data, Narrator achieves better performance on all tasks compared with BART-CNN-GEN and BART-CRD3-GEN, demonstrating that DIANA is a more helpful resource for dialogue comprehension.

\begin{table}[t]
		\centering
		\small
		\setlength\tabcolsep{10pt}
\begin{tabular}{l|cc}
 \toprule
Question Type & BART & Narrator \\
\midrule
Paraphrase+Matching & $58.4$ & $66.1\enspace (+7.7)$ \\
Reasoning & $42.2$ & $46.2\enspace (+4.0)$ \\
\quad Summary & $51.1$ & $53.4\enspace (+2.3)$ \\
\quad Logic & $43.8$ & $48.2\enspace (+4.4)$ \\
\quad Commonsense & $37.8$ & $41.9\enspace (+4.1)$ \\
\quad Arithmetic & $23.8$ & $23.8\enspace (+0.0)$ \\
 \bottomrule
\end{tabular}

		\caption{\label{tab::dream}Accuracy by question types on DREAM.}
		\vspace{-8pt}
\end{table}

We further analyze what types of knowledge are enhanced during pre-training. To this end, we test Narrator on a subset of the DREAM test set, which includes annotated knowledge types released along with the DREAM dataset. As shown in Table \ref{tab::dream}, compared with the vanilla BART, Narrator achieves better performance on all knowledge types except Arithmetic, which is not covered in DIANA. The performance gain indicates that the narrative pre-training contributes the most to the knowledge related to paraphrasing and matching. It also benefits from other knowledge types that require various reasoning abilities such as commonsense reasoning and logic reasoning. %

\section{Conclusion}
We propose a \emph{learning-by-narrating} strategy to pre-train a zero-shot dialogue comprehension model. We first construct a dialogue-narrative dataset named DIANA, which contains 243K (dialogue, narrative) pairs obtained by automatically aligning movie subtitles with their corresponding synopses. We then pre-train a dialogue comprehension model based on DIANA and evaluate its performance on four downstream tasks that require dialogue comprehension abilities. Experiments show that our model outperforms strong pre-trained baselines, demonstrating that the learning-by-narrating strategy is a promising direction for dialogue comprehension. We also hope that DIANA will promote future research in related areas.

\bibliography{custom}

\begin{thebibliography}{26}
\expandafter\ifx\csname natexlab\endcsname\relax\def\natexlab#1{#1}\fi

\bibitem[{Afantenos et~al.(2015)Afantenos, Kow, Asher, and
  Perret}]{afantenos-2015-discourse}
Stergos Afantenos, Eric Kow, Nicholas Asher, and J{\'e}r{\'e}my Perret. 2015.
\newblock \href {https://doi.org/10.18653/v1/D15-1109} {Discourse parsing for
  multi-party chat dialogues}.
\newblock In \emph{Proceedings of the 2015 Conference on Empirical Methods in
  Natural Language Processing}, pages 928--937, Lisbon, Portugal. Association
  for Computational Linguistics.

\bibitem[{Arabshahi et~al.(2021)Arabshahi, Lee, Gawarecki, Mazaitis, Azaria,
  and Mitchell}]{arabshahi2021conversational}
Forough Arabshahi, Jennifer Lee, Mikayla Gawarecki, Kathryn Mazaitis, Amos
  Azaria, and Tom Mitchell. 2021.
\newblock Conversational neuro-symbolic commonsense reasoning.
\newblock In \emph{Proceedings of the AAAI Conference on Artificial
  Intelligence}, volume~35, pages 4902--4911.

\bibitem[{Chapuis et~al.(2020)Chapuis, Colombo, Manica, Labeau, and
  Clavel}]{chapuis2020hierarchical}
Emile Chapuis, Pierre Colombo, Matteo Manica, Matthieu Labeau, and Chlo{\'e}
  Clavel. 2020.
\newblock \href {https://doi.org/10.18653/v1/2020.findings-emnlp.239}
  {Hierarchical pre-training for sequence labelling in spoken dialog}.
\newblock In \emph{Findings of the Association for Computational Linguistics:
  EMNLP 2020}, pages 2636--2648, Online. Association for Computational
  Linguistics.

\bibitem[{Chen et~al.(2021)Chen, Choi, and Durrett}]{chen-etal-2021-nli-models}
Jifan Chen, Eunsol Choi, and Greg Durrett. 2021.
\newblock \href {https://aclanthology.org/2021.findings-emnlp.324} {Can {NLI}
  models verify {QA} systems{'} predictions?}
\newblock In \emph{Findings of the Association for Computational Linguistics:
  EMNLP 2021}, pages 3841--3854, Punta Cana, Dominican Republic.

\bibitem[{Cui et~al.(2020)Cui, Wu, Liu, Zhang, and Zhou}]{cui-2020-mutual}
Leyang Cui, Yu~Wu, Shujie Liu, Yue Zhang, and Ming Zhou. 2020.
\newblock \href {https://doi.org/10.18653/v1/2020.acl-main.130} {{M}u{T}ual: A
  dataset for multi-turn dialogue reasoning}.
\newblock In \emph{Proceedings of the 58th Annual Meeting of the Association
  for Computational Linguistics}, pages 1406--1416, Online. Association for
  Computational Linguistics.

\bibitem[{Devlin et~al.(2019)Devlin, Chang, Lee, and
  Toutanova}]{devlin-2019-bert}
Jacob Devlin, Ming-Wei Chang, Kenton Lee, and Kristina Toutanova. 2019.
\newblock \href {https://doi.org/10.18653/v1/N19-1423} {{BERT}: Pre-training of
  deep bidirectional transformers for language understanding}.
\newblock In \emph{Proceedings of the 2019 Conference of the North {A}merican
  Chapter of the Association for Computational Linguistics: Human Language
  Technologies, Volume 1 (Long and Short Papers)}, pages 4171--4186,
  Minneapolis, Minnesota. Association for Computational Linguistics.

\bibitem[{Falke et~al.(2020)Falke, Boese, Sorokin, Tirkaz, and
  Lehnen}]{falke-etal-2020-leveraging}
Tobias Falke, Markus Boese, Daniil Sorokin, Caglar Tirkaz, and Patrick Lehnen.
  2020.
\newblock \href {https://doi.org/10.18653/v1/2020.coling-industry.3}
  {Leveraging user paraphrasing behavior in dialog systems to automatically
  collect annotations for long-tail utterances}.
\newblock In \emph{Proceedings of the 28th International Conference on
  Computational Linguistics: Industry Track}, pages 21--32, Online.
  International Committee on Computational Linguistics.

\bibitem[{Gliwa et~al.(2019)Gliwa, Mochol, Biesek, and Wawer}]{gliwa2019samsum}
Bogdan Gliwa, Iwona Mochol, Maciej Biesek, and Aleksander Wawer. 2019.
\newblock \href {https://doi.org/10.18653/v1/D19-5409} {{SAMS}um corpus: A
  human-annotated dialogue dataset for abstractive summarization}.
\newblock In \emph{Proceedings of the 2nd Workshop on New Frontiers in
  Summarization}, pages 70--79, Hong Kong, China. Association for Computational
  Linguistics.

\bibitem[{Grusky et~al.(2018)Grusky, Naaman, and Artzi}]{grusky2018newsroom}
Max Grusky, Mor Naaman, and Yoav Artzi. 2018.
\newblock \href {https://doi.org/10.18653/v1/N18-1065} {{N}ewsroom: A dataset
  of 1.3 million summaries with diverse extractive strategies}.
\newblock In \emph{Proceedings of the 2018 Conference of the North {A}merican
  Chapter of the Association for Computational Linguistics: Human Language
  Technologies, Volume 1 (Long Papers)}, pages 708--719, New Orleans,
  Louisiana. Association for Computational Linguistics.

\bibitem[{Huang et~al.(2020)Huang, Xiong, Rao, Wang, and
  Lin}]{huang2020movienet}
Qingqiu Huang, Yu~Xiong, Anyi Rao, Jiaze Wang, and Dahua Lin. 2020.
\newblock Movienet: A holistic dataset for movie understanding.
\newblock In \emph{Computer Vision--ECCV 2020: 16th European Conference,
  Glasgow, UK, August 23--28, 2020, Proceedings, Part IV 16}, pages 709--727.
  Springer.

\bibitem[{Jin et~al.(2020)Jin, Gao, Kao, Chung, and
  Hakkani{-}T{\"{u}}r}]{jin2019mmm}
Di~Jin, Shuyang Gao, Jiun{-}Yu Kao, Tagyoung Chung, and Dilek
  Hakkani{-}T{\"{u}}r. 2020.
\newblock \href {https://aaai.org/ojs/index.php/AAAI/article/view/6310} {{MMM:}
  multi-stage multi-task learning for multi-choice reading comprehension}.
\newblock In \emph{The Thirty-Fourth {AAAI} Conference on Artificial
  Intelligence, {AAAI} 2020, The Thirty-Second Innovative Applications of
  Artificial Intelligence Conference, {IAAI} 2020, The Tenth {AAAI} Symposium
  on Educational Advances in Artificial Intelligence, {EAAI} 2020, New York,
  NY, USA, February 7-12, 2020}, pages 8010--8017. {AAAI} Press.

\bibitem[{Lei et~al.(2020)Lei, Yu, Berg, and Bansal}]{lei2020more}
Jie Lei, Licheng Yu, Tamara Berg, and Mohit Bansal. 2020.
\newblock \href {https://doi.org/10.18653/v1/2020.emnlp-main.706} {What is more
  likely to happen next? video-and-language future event prediction}.
\newblock In \emph{Proceedings of the 2020 Conference on Empirical Methods in
  Natural Language Processing (EMNLP)}, pages 8769--8784, Online. Association
  for Computational Linguistics.

\bibitem[{Lewis et~al.(2020)Lewis, Liu, Goyal, Ghazvininejad, Mohamed, Levy,
  Stoyanov, and Zettlemoyer}]{lewis2019bart}
Mike Lewis, Yinhan Liu, Naman Goyal, Marjan Ghazvininejad, Abdelrahman Mohamed,
  Omer Levy, Veselin Stoyanov, and Luke Zettlemoyer. 2020.
\newblock \href {https://doi.org/10.18653/v1/2020.acl-main.703} {{BART}:
  Denoising sequence-to-sequence pre-training for natural language generation,
  translation, and comprehension}.
\newblock In \emph{Proceedings of the 58th Annual Meeting of the Association
  for Computational Linguistics}, pages 7871--7880, Online. Association for
  Computational Linguistics.

\bibitem[{Lison et~al.(2018)Lison, Tiedemann, and
  Kouylekov}]{lison2018opensubtitles2018}
Pierre Lison, J{\"o}rg Tiedemann, and Milen Kouylekov. 2018.
\newblock \href {https://aclanthology.org/L18-1275} {{O}pen{S}ubtitles2018:
  Statistical rescoring of sentence alignments in large, noisy parallel
  corpora}.
\newblock In \emph{Proceedings of the Eleventh International Conference on
  Language Resources and Evaluation ({LREC} 2018)}, Miyazaki, Japan. European
  Language Resources Association (ELRA).

\bibitem[{Liu et~al.(2021)Liu, Feng, Wang, Song, Ren, and Zhang}]{liu2021graph}
Yongkang Liu, Shi Feng, Daling Wang, Kaisong Song, Feiliang Ren, and Yifei
  Zhang. 2021.
\newblock A graph reasoning network for multi-turn response selection via
  customized pre-training.
\newblock In \emph{Proceedings of the AAAI Conference on Artificial
  Intelligence}, volume~35, pages 13433--13442.

\bibitem[{Ma et~al.(2018)Ma, Jurczyk, and Choi}]{ma-etal-2018-challenging}
Kaixin Ma, Tomasz Jurczyk, and Jinho~D. Choi. 2018.
\newblock \href {https://doi.org/10.18653/v1/N18-1185} {Challenging reading
  comprehension on daily conversation: Passage completion on multiparty
  dialog}.
\newblock In \emph{Proceedings of the 2018 Conference of the North {A}merican
  Chapter of the Association for Computational Linguistics: Human Language
  Technologies, Volume 1 (Long Papers)}, pages 2039--2048, New Orleans,
  Louisiana. Association for Computational Linguistics.

\bibitem[{Qiu et~al.(2021)Qiu, Zhang, and Zhou}]{qiu-2021-different}
Yao Qiu, Jinchao Zhang, and Jie Zhou. 2021.
\newblock \href {https://aclanthology.org/2021.emnlp-main.178} {Different
  strokes for different folks: Investigating appropriate further pre-training
  approaches for diverse dialogue tasks}.
\newblock In \emph{Proceedings of the 2021 Conference on Empirical Methods in
  Natural Language Processing}, pages 2318--2327, Online and Punta Cana,
  Dominican Republic.

\bibitem[{Radford et~al.(2019)Radford, Wu, Child, Luan, Amodei, Sutskever
  et~al.}]{radford2019language}
Alec Radford, Jeffrey Wu, Rewon Child, David Luan, Dario Amodei, Ilya
  Sutskever, et~al. 2019.
\newblock \href
  {https://d4mucfpksywv.cloudfront.net/better-language-models/language_models_are_unsupervised_multitask_learners.pdf}
  {Language models are unsupervised multitask learners}.
\newblock \emph{OpenAI blog}, 1(8):9.

\bibitem[{Rameshkumar and Bailey(2020)}]{crd3}
Revanth Rameshkumar and Peter Bailey. 2020.
\newblock \href {https://doi.org/10.18653/v1/2020.acl-main.459} {Storytelling
  with dialogue: {A} {Critical Role Dungeons and Dragons Dataset}}.
\newblock In \emph{Proceedings of the 58th Annual Meeting of the Association
  for Computational Linguistics}, pages 5121--5134, Online. Association for
  Computational Linguistics.

\bibitem[{See et~al.(2017)See, Liu, and Manning}]{see-2017-get}
Abigail See, Peter~J. Liu, and Christopher~D. Manning. 2017.
\newblock \href {https://doi.org/10.18653/v1/P17-1099} {Get to the point:
  Summarization with pointer-generator networks}.
\newblock In \emph{Proceedings of the 55th Annual Meeting of the Association
  for Computational Linguistics (Volume 1: Long Papers)}, pages 1073--1083,
  Vancouver, Canada. Association for Computational Linguistics.

\bibitem[{Sun et~al.(2019)Sun, Yu, Chen, Yu, Choi, and Cardie}]{sun-2019-dream}
Kai Sun, Dian Yu, Jianshu Chen, Dong Yu, Yejin Choi, and Claire Cardie. 2019.
\newblock \href {https://doi.org/10.1162/tacl_a_00264} {{DREAM}: A challenge
  data set and models for dialogue-based reading comprehension}.
\newblock \emph{Transactions of the Association for Computational Linguistics},
  7:217--231.

\bibitem[{Tapaswi et~al.(2015)Tapaswi, B{\"a}uml, and
  Stiefelhagen}]{tapaswi2015aligning}
Makarand Tapaswi, Martin B{\"a}uml, and Rainer Stiefelhagen. 2015.
\newblock \href
  {http://www.cs.toronto.edu/~makarand/papers/IJMIR_plot-retrieval.pdf}
  {Aligning plot synopses to videos for story-based retrieval}.
\newblock \emph{International Journal of Multimedia Information Retrieval},
  4(1):3--16.

\bibitem[{Wu et~al.(2020)Wu, Hoi, Socher, and Xiong}]{wu-etal-2020-tod}
Chien-Sheng Wu, Steven~C.H. Hoi, Richard Socher, and Caiming Xiong. 2020.
\newblock \href {https://doi.org/10.18653/v1/2020.emnlp-main.66} {{TOD}-{BERT}:
  Pre-trained natural language understanding for task-oriented dialogue}.
\newblock In \emph{Proceedings of the 2020 Conference on Empirical Methods in
  Natural Language Processing (EMNLP)}, pages 917--929, Online. Association for
  Computational Linguistics.

\bibitem[{Yin et~al.(2021)Yin, Radev, and Xiong}]{yin-etal-2021-docnli}
Wenpeng Yin, Dragomir Radev, and Caiming Xiong. 2021.
\newblock \href {https://doi.org/10.18653/v1/2021.findings-acl.435}
  {{D}oc{NLI}: A large-scale dataset for document-level natural language
  inference}.
\newblock In \emph{Findings of the Association for Computational Linguistics:
  ACL-IJCNLP 2021}, pages 4913--4922, Online. Association for Computational
  Linguistics.

\bibitem[{Zhang and Zhou(2019)}]{zhang2019automatically}
Leilan Zhang and Qiang Zhou. 2019.
\newblock \href {https://ieeexplore.ieee.org/abstract/document/9023129/}
  {Automatically annotate tv series subtitles for dialogue corpus
  construction}.
\newblock In \emph{2019 Asia-Pacific Signal and Information Processing
  Association Annual Summit and Conference (APSIPA ASC)}, pages 1029--1035.
  IEEE.

\bibitem[{Zhang and Zhao(2021)}]{zhang-zhao-2021-structural}
Zhuosheng Zhang and Hai Zhao. 2021.
\newblock \href {https://doi.org/10.18653/v1/2021.acl-long.399} {Structural
  pre-training for dialogue comprehension}.
\newblock In \emph{Proceedings of the 59th Annual Meeting of the Association
  for Computational Linguistics and the 11th International Joint Conference on
  Natural Language Processing (Volume 1: Long Papers)}, pages 5134--5145,
  Online. Association for Computational Linguistics.

\end{thebibliography}
\bibliographystyle{acl_natbib}

\clearpage
\newpage
\appendix

\end{document}